%% file: main.tex
\documentclass{article}

\usepackage{microtype}
\usepackage{graphicx}
\usepackage{subcaption}
\usepackage{booktabs} %

\usepackage{hyperref}
\usepackage{multirow}
\usepackage{adjustbox}
\usepackage{amsmath,bm}
\usepackage[table]{xcolor}
\definecolor{rowgray}{gray}{0.9}

\usepackage[preprint]{icml2026}

\usepackage{amsmath}
\usepackage{amssymb}
\usepackage{mathtools}
\usepackage{amsthm}

\usepackage[capitalize,noabbrev]{cleveref}

\theoremstyle{plain}

\theoremstyle{definition}

\theoremstyle{remark}

\usepackage[textsize=tiny]{todonotes}

\icmltitlerunning{}

\begin{document}

\twocolumn[
  \icmltitle{Efficient Training of Diffusion Mixture-of-Experts Models: A Practical Recipe}

  \icmlsetsymbol{equal}{*}

  \begin{icmlauthorlist}
    \icmlauthor{Yahui Liu, Yang Yue, Jingyuan Zhang, Chenxi Sun, Yang Zhou, Wencong Zeng, Ruiming Tang, Guorui Zhou$^\dagger$}{yyy}
  \end{icmlauthorlist}

  \icmlaffiliation{yyy}{Kuaishou Technology, Beijing, China}

  \icmlcorrespondingauthor{Guorui Zhou}{zhouguorui@kuaishou.com}

  \icmlkeywords{Machine Learning, ICML}

  \vskip 0.3in
]

\printAffiliationsAndNotice{}  %

\input{sections/0-abstract}

\input{sections/1-instro}

\input{sections/2-related-work}

\input{sections/3-moe-design}

\input{sections/4-experiments}

\input{sections/5-conclusion}

\bibliography{example_paper}
\bibliographystyle{icml2026}

\input{sections/appendix}

\end{document}

%% file: sections/0-abstract.tex
\begin{abstract}
  Recent efforts on Diffusion Mixture-of-Experts (MoE) models have primarily focused on developing more sophisticated routing mechanisms. However, we observe that the underlying architectural configuration space remains markedly under-explored. 
  Inspired by the MoE design paradigms established in large language models (LLMs), we identify a set of crucial architectural factors for building effective Diffusion MoE models--including DeepSeek-style expert modules, alternative intermediate widths, varying expert counts, and enhanced attention positional encodings. 
  Our systematic study reveals that carefully tuning these configurations is essential for unlocking the full potential of Diffusion MoE models, often yielding gains that exceed those achieved by routing innovations alone. 
  Through extensive experiments, we present novel architectures that can be efficiently applied to both latent and pixel-space diffusion frameworks, which provide a practical and efficient training recipe that enables Diffusion MoE models to surpass strong baselines while using equal or fewer activated parameters. All code and models are publicly available at: \url{https://github.com/yhlleo/EfficientMoE}.
\end{abstract}

%% file: sections/1-instro.tex
\section{Introduction}
\label{sec:intro}

The Diffusion Transformer (DiT)~\cite{peebles2023scalable} architecture represents a novel approach to diffusion-based visual generation, combining the power of transformers~\cite{vaswani2017attention} with the flexibility and efficiency of diffusion models~\cite{ho2020denoising,song2020score}. DiT leverages the inherent scalability and expressiveness of transformers, making it particularly well-suited for generating high-quality, high-fidelity images from random noise~\cite{esser2024scaling,yang2024cogvideox}.

Inspired by the success of Mixture-of-Experts (MoE) models in the large language model (LLM) community~\cite{dai2024deepseekmoe,liu2024deepseekv2,liu2024deepseekv3,yang2025qwen3,team2025kimi}, there has been increasing interest in integrating MoE into diffusion architectures, espcially those based on DiT framework. In LLMs, MoE allows the model to scale its capacity dramatically while activating only a small subset of parameters per token. This efficiency makes MoE an attractive direction for diffusion models as they also grow larger and more capable. 

However, recent efforts~\cite{fei2024scaling,sun2024ec} to integrate it into DiT models
have not yielded the significant gains observed in LLMs. Prior works~\cite{shi2025diffmoe,yuan2025expert,wei2025routing} often attribute this gap to the fundamental differences between text tokens and visual tokens. As a result, research efforts have tended to center around designing routing networks that can intelligently select experts based on factors such as noise level or input conditions. 
In contrast, our investigation reveals that the core limitation lies elsewhere: \textit{the architectural design space of MoE for diffusion models remains significantly under-explored}, and this under-exploration is the primary reason behind the performance gap. Notably, by taking established LLM MoE architecture as our starting point and conducting a broad, multi-dimensional investigation, we identify a set of design principles that lead to efficient and effective Diffusion MoE models.

Finally, we distill several key findings: 
\begin{itemize}
    \item \textit{MoE modules that have proven effective in LLMs remain highly effective in diffusion models}. For example, the DeepSeek-V3~\cite{liu2024deepseekv3} MoE block, with minor modifications, already yields strong performance in diffusion tasks. 
    \item \emph{The MoE MLP intermediate-size scaling can follow similar principles to those in LLM MoE models.} Setting the scaling factor to below 4.0 reduces both total parameters and activated parameters. When combined with an appropriate number of active experts, this reduction does not degrade the performance of Diffusion MoE models.
    \item \emph{Using shared experts tends to improve training stability and convergence.} Our observations indicate that incorporating shared experts provides a beneficial regularizing effect.
    \item  \emph{Incorporating 2D RoPE into the attention layers significantly improves training.} Adding explicit 2D positional information enhances spatial reasoning and contributes to more stable and effective model optimization. 
\end{itemize}

We present our DSMoE and JiTMoE, the novel MoE architectures that applies the above principles to both latent~\cite{rombach2022high} and pixel-space~\cite{li2025back} diffusion frameworks of various scales (\emph{e.g.}, from 100M to 3B parameters) and observe consistent performance gains on the Flow Matching~\cite{lipman2022flow,esser2024scaling} paradigm. Especially, our ``XL'' model achieves substantial improvements over prior methods with comparable activation parameters on latent diffusion framework. 
Our 3B variant establishes a new state of the art among diffusion-based models, attaining a best FID of 2.38 on class-conditional ImageNet~\cite{deng2009imagenet} 256$\times$256 with only 700K training steps. Notably, this matches the performance of DiffMoE~\cite{chengdiff}, which reports a 2.30 FID but requires an order-of-magnitude more training (7000K steps).
We release all code, model configurations, and pretrained weights to the community. To the best of our knowledge, this is the first open-source model in this direction, enabling future researchers to conduct deeper and more systematic exploration.

%% file: sections/2-related-work.tex
\section{Related Work}
\label{sec:related-work}

\noindent\textbf{Diffusion Models.} Diffusion models~\cite{ho2020denoising,song2020score} have emerged as a dominant and highly effective paradigm for high-quality visual generation. Recent work~\cite{chen2023pixart,ma2024sit,hatamizadeh2024diffit,chu2024visionllama,esser2024scaling,wei2025dreamrelation} has achieved impressive scalability and generative quality by adopting the Diffusion Transformers (DiT)~\cite{peebles2023scalable} on the latent diffusion~\cite{rombach2022high} framework. Such DiT models are usually trained with two primary paradigms: Denoising Diffusion Probabilistic Models (DDPM) trained via score matching~\cite{ho2020denoising,song2020score} and Rectified Flow (RF)~\cite{liu2022flow} approaches optimized through flow-matching~\cite{lipman2022flow,esser2024scaling,ma2024sit}. Recently, JiT~\cite{li2025back}, introduced a new diffusion framework operating in the pixel space, exhibits remarkably strong performance.  In this work, we implement our MoE designs on both the latent and pixel-space diffusion frameworks using only the RF paradigm.

\noindent\textbf{Mixture of Experts.} Mixture-of-Experts (MoE)~\cite{jacobs1991adaptive,shazeer2017outrageously,lepikhin2020gshard} is designed to increase model capacity by routing each input to a small subset of specialized experts, enabling efficient sparse computation. Recently, such approach has demonstrated remarkable success in Large Language Models (LLMs)~\cite{dai2024deepseekmoe,liu2024deepseekv3,yang2025qwen3,team2025kimi}. However, adapting MoE to DiT architecture faces some limitations~\cite{sehwag2025stretching,chengdiff,fei2024scaling,sun2024ec,shi2025diffmoe,wei2025routing} to achieve appealing performance. A common line of analysis attributes this gap to fundamental differences between language and vision tokens. Consequently, many works explore diverse routing strategies that adapt expert selection based on noise level and input conditions. Distinct from prior work, we are the first to systematically explore the architecture design space of MoE and demonstrate that improved MoE structures alone offer significant performance potential.

%% file: sections/3-moe-design.tex
\section{Architecture Design}
\label{sec:moe-design}

\begin{table*}[ht]
  \centering
\caption{DSMoE and JiTMoE Model Configurations. DSMoE is designed for latent diffusion models (LDMs), whereas JiTMoE is tailored for pixel-space diffusion models. Hyperparameter settings and computational specifications for class-conditional models. ``S1E16A2"" denotes that a shared expert shared by all tokens, 16 experts used for routing, and 2 experts activated for each token. 
}
  {
    \begin{tabular}{lccccccc}\toprule
    Model 
    & \#Activated  
    & \#Total 
    & \#Blocks 
    & \#Hidden 
    & \#Intermediate
    & \#Head 
    & \#Experts
    \\
    Config &
    Params. & 
    Params. & 
    L & 
    dim. D &
    size S &
    n
    \\ \midrule
    DSMoE-S-E16 & 33M   & 92M    & 10  & 384  & 768 & 6  & S1E16A2 \\
    DSMoE-S-E48 & 30M   & 66M    & 12  & 384  & 128 & 6  & S1E48A5 \\
    DSMoE-B-E16 & 132M  & 368M    & 10  & 768  & 1536 & 12 & S1E16A2 \\
    DSMoE-B-E48 & 118M  & 263M    & 12  & 768  & 256 & 12 & S1E48A5 \\
    DSMoE-L-E16 & 465M  & 1.304B  & 20  & 1024 & 2048 & 16 & S1E16A2 \\
    DSMoE-L-E48 & 436M  & 1.112B  & 24  & 1024 & 448 & 16 & S1E48A5 \\
    DSMoE-3B-E16 & 965M  & 2.958B  & 30  & 1152 & 2880 & 16 & S1E16A2 \\ \midrule
    JiTMoE-B/16-E16 & 133M & 369M & 10 & 768 & 1536 & 12 & S1E16A2 \\
    JiTMoE-L/16-E16 &  465M & 1.306B & 20 & 1024 & 2048 & 16 & S1E16A2 \\
    \bottomrule
    \end{tabular}%
    }
  \label{tab:model_arch}%
\end{table*}%

\noindent\textbf{Diffusion Models.} Diffusion models are generative models that learn data distributions by reversing a forward noising process.
The continuous-time forward process can be formulated as $\bm{x}_t = \alpha_t \bm{x}_0 + \sigma_t \bm{\epsilon}$, with $t \in \mathcal{U}(0, 1)$ and $\bm{\epsilon} \sim \mathcal{N}(0, \mathbf{{I}})$.
$\alpha_t$ and $\sigma_t$ are monotonically decreasing and increasing functions of $t$, respectively.
For the reverse process,
a denoising network $\mathcal{F}_\theta$ is trained to predict the target $\bm{y}$ at each timestep $t$, conditioned on $\bm{c}$ (\textit{e.g.}, class labels or text prompts): 
\begin{align}
    \mathcal{L} = \mathbb{E}_{\bm{x}_0, \bm{c}, \bm{\epsilon}, t}\left[\left\| \bm{y} - \mathcal{F}_\theta\left(\bm{x}_t, \bm{c}, t\right)\right\|_2^2\right],
\label{eq:diffusion_loss}
\end{align}
where the training target $\bm{y}$ can be the Gaussian noise $\bm{\epsilon}$ for DDPM models~\citep{ho2020denoising}, or the vector field $(\bm{\epsilon} - \bm{x}_0)$ for Rectified Flow models~\citep{liu2022flow}. In this work, our discussions focus on Rectified Flow.

\noindent\textbf{MoE Module.} We employ the MoE module proposed in DeepSeekMoE~\cite{liu2024deepseekv3,dai2024deepseekmoe} as our Feed-Forward Networks (FFNs). Different from traditional MoE architectures, DeepSeekMoE uses finer-grained experts and isolates some experts as shared ones. Let $\bm{u}_t$ denote the FFN input of the $t$-th token, the FFN output $\bm{h}_t$ is computed as follows:
\begin{align}
    \bm{h}_{t} & = \bm{u}_{t} + \sum_{i=1}^{N_{s}} {\operatorname{FFN}^{(s)}_{i}\left( \bm{u}_{t} \right)} + \sum_{i=1}^{N_r} {g_{i,t} \operatorname{FFN}^{(r)}_{i}\left( \bm{u}_{t} \right)}, \\
    g_{i,t} & = \frac{g^{\prime}_{i,t}}{\sum_{j=1}^{N_r} g^{\prime}_{j,t}}, \\
    g^{\prime}_{i,t} & = \begin{cases} 
    s_{i,t}, & s_{i,t} \in \operatorname{Topk} (\{ s_{j, t} | 1 \leq j \leq N_r \}, K_{r}), \\
    0, & \text{otherwise}, 
    \end{cases} \label{eq3} \\
    s_{i,t} & = \operatorname{Sigmoid} \left( {\bm{u}_{t}}^{T} \bm{e}_{i} \right),
\end{align}
where $N_{s}$ and $N_r$ denote the numbers of shared experts and routed experts, respectively; 
$\operatorname{FFN}^{(s)}_{i}(\cdot)$ and $\operatorname{FFN}^{(r)}_{i}(\cdot)$ denote the $i$-th shared expert and the $i$-th routed expert, respectively; 
$K_{r}$ denotes the number of activated routed experts; 
$g_{i,t}$ is the gating value for the $i$-th expert; 
$s_{i,t}$ is the token-to-expert affinity; 
$\mathbf{e}_{i}$ is the centroid vector of the $i$-th routed expert; 
and $\operatorname{Topk}(\cdot, K)$ denotes the set comprising $K$ highest scores among the affinity scores calculated for the $t$-th token and all routed experts.
In our experiments, we set $N_{s}=1$ and $K_{r}$ to either 2 or 5, depending on the model configuration. 

To achieve a better trade-off between load balance
and model performance, we follow the auxiliary-loss-free load balancing strategy~\cite{wang2024auxiliary}. In this strategy, a bias term $b_i$ is introduced for each expert and added to the corresponding affinity score $s_{i,t}$ to determine the Top-K routing. Thus, the Eq.~(\ref{eq3}) is adjusted to:

\begin{align}
    g^{\prime}_{i,t} & = \begin{cases} 
    s_{i,t}, & s_{i,t} + b_i \in \operatorname{Topk} (\{ s_{j, t} + b_j | 1 \leq j \leq N_r \}, K_{r}), \\
    0, & \text{otherwise}.
    \end{cases}
\end{align}

Note that the bias term is only used for routing.
The gating value, which will be multiplied with the FFN output, is still derived from the original affinity score $s_{i,t}$.

\noindent\textbf{Position Encoding.} We replace absolute position embeddings with 2D rotary position embeddings (RoPE) applied to queries and keys in both self-attention and cross-attention. Standard RoPE~\cite{su2024roformer} rotates each query/key dimension by a phase proportional to its position, so relative offsets are captured by phase differences. Let $\mathbf{R}_{\Theta}$ be the rotary matrix defined by a frequency vector $\Theta$; the RoPE attention between token $i$ and $j$ is
\begin{align}
    \mathbf{A}_{i,j}^{\text{RoPE}} &= \boldsymbol{q}_i^\top \mathbf{R}_{\Theta,i}^\top \mathbf{R}_{\Theta,j} \boldsymbol{k}_j, \\
    \mathbf{R}_{\Theta,j-i} &= \mathbf{R}_{\Theta,i}^\top \mathbf{R}_{\Theta,j},
\end{align}
so the product depends on the relative displacement $j-i$. For 2D inputs (\textit{e.g.}, images and their latent counterparts), flattening tokens discards the separable row/column structure. 2D RoPE restores this by assigning independent rotary phases to height and width indices and interleaving them, so the resulting phase encodes both axes. This preserves relative geometry on 2D grids, adds no parameters, and keeps attention equivariant to translations/rotations in the rotary subspace.

%% file: sections/4-experiments.tex
\section{Experimental Results}
\label{sec:experiments}

\begin{table*}[t]
  \centering
\caption{Quantitative comparison with DiffMoE~\cite{shi2025diffmoe} on various model sizes under Rectified Flow. All models are trained on ImageNet 256x256 with 700K steps (batch size 512). Note that DiffMoE-L-E16 (CFG=1.0) reports FID and IS scores 14.41 and 88.19, respectively. The version of DiffMoE-L-E16 we reproduced performs even better.  
}
  {
    \begin{tabular}{lccccccc}\toprule
    \multirow{2}{*}{Model} & 
    \#Activated  & \#Total
    & \multicolumn{2}{c}{CFG = 1.0}
    & \multicolumn{2}{c}{CFG = 1.5} 
    \\ \cmidrule(lr){4-5} \cmidrule(lr){6-7}
    & Params. & Params. & FID ($\downarrow$) & IS ($\uparrow$) & FID ($\downarrow$) & IS ($\uparrow$)
    \\ \midrule
    DiffMoE-S-E16 & 32M & 139M & 41.02 & 37.53 & 15.47 & 94.04 \\
    \rowcolor{rowgray}
    DSMoE-S-E16 & 33M & 92M & 39.84 & 38.63 & 14.53 & 97.55 \\
    \rowcolor{rowgray}
    DSMoE-S-E48 & 30M & 66M & 40.20 & 38.09 & 14.81 & 96.51 \\
    \midrule
    DiffMoE-B-E16 & 130M & 555M & 20.83 & 70.26 & 4.87 & 183.43 \\
    \rowcolor{rowgray}
    DSMoE-B-E16 & 132M & 368M & 20.33 & 71.42 & 4.50 & 186.79 \\
    \rowcolor{rowgray}
    DSMoE-B-E48 & 118M & 263M & 19.46 & 72.69 & 4.27 & 191.03 \\
    \midrule
    DiffMoE-L-E16 & 458M & 1.982B & 11.16 & 107.74 & 2.84 & 256.57 \\
    \rowcolor{rowgray}
    DSMoE-L-E16 & 465M & 1.304B & 9.80 & 115.45 & 2.59 & 272.55 \\
    \rowcolor{rowgray}
    DSMoE-L-E48 & 436M & 1.112B & 9.19 & 118.52 & 2.55 & 278.35 \\
    \midrule
    \rowcolor{rowgray}
    DSMoE-3B-E16 & 965M & 2.958B &7.52 & 135.29 & 2.38 & 304.93 \\
    \bottomrule
    \end{tabular}%
    }
  \label{tab:compared_diffmoe}%
\end{table*}%

\begin{table*}[ht]
  \centering
\caption{Quantitative comparison with Dense-DiT~\cite{peebles2023scalable} and ProMoE~\cite{wei2025routing} on various model sizes under Rectified Flow. All models are trained on ImageNet 256x256 with 500K steps (batch size 512). 
}
  {
    \begin{tabular}{lccccccc}\toprule
    \multirow{2}{*}{Model} &
    \#Activated  & \#Total
    & \multicolumn{2}{c}{CFG = 1.0}
    & \multicolumn{2}{c}{CFG = 1.5} 
    \\ \cmidrule(lr){4-5} \cmidrule(lr){6-7}
    & Params. & Params. & FID ($\downarrow$) & IS ($\uparrow$) & FID ($\downarrow$) & IS ($\uparrow$)
    \\ \midrule
    Dense-DiT-B & 130M & 130M & 30.61 & 49.89 &  9.02 &  131.13 \\
    ProMoE-B & 130M & 300M & 24.44 & 60.38 & 6.39 & 154.21 \\
    \rowcolor{rowgray}
    DSMoE-B-E16 & 132M & 368M & 22.46 & 64.57 & 5.38 & 170.22 \\
    \rowcolor{rowgray}
    DSMoE-B-E48 & 118M & 263M & 21.56 & 66.48 & 5.14 & 172.80 \\
    \midrule
    Dense-DiT-L & 458M & 458M & 15.44 & 84.20 &  3.56 & 209.03 \\
    ProMoE-L & 458M & 1.063B & 11.61 & 100.82 & 2.79  & 244.21 \\
    \rowcolor{rowgray}
    DSMoE-L-E16 &  465M & 1.304B & 10.71 & 107.34 & 2.78 & 258.09 \\
    \rowcolor{rowgray}
    DSMoE-L-E48 & 436M & 1.112B & 9.87 & 111.54 & 2.72 & 263.13 \\
    \bottomrule
    \end{tabular}%
    }
  \label{tab:compared_promoe}%
\end{table*}%

\subsection{Experimental Setup}

\noindent\textbf{Baseline and Model architecture.} We compare with existing state-of-the-art diffusion MoE models with latent diffusion framework by flow matching~\cite{lipman2022flow}, including DiffMoE~\cite{chengdiff} and ProMoE~\cite{wei2025routing}. We compare with the dense pixel-space diffusion framework JiT~\cite{li2025back} by flow matching. Our models are named as: [Model]-[Size]-[\# Experts]. Similar to previous work, we replace even FFN layers with MoE layers containing $N$ identical FFN components~\cite{lepikhin2020gshard}, and replace the absolute position embeddings in the original DiT architeture with 2D RoPE. Details of architecture are presented in Table~\ref{tab:model_arch}. All these class-conditional diffusion MoE models are trained at 256$\times$256 image resolution on the ImageNet dataset~\cite{russakovsky2015imagenet}, which contains 1,281,167 training images. 

\noindent\textbf{Evaluation.} We evaluate our DSMoE and JiTMoE models through both quantitative and qualitative metrics. Quantitatively, we evaluate DSMoE, DiffMoE and ProMoE used FID50K~\cite{heusel2017gans} with DDPM/Euler steps for class-conditional generation. We evaluate JiTMoE by following the configuration of JiT by setting ODE steps to 50 and with CFG interval~\cite{kynkaanniemi2024applying}. Notably, we keep the original training and evaluation configurations/protocols of both latent-diffusion and pixel-space diffusion frameworks unchanged to ensure a fair comparison. Similar to previous work~\cite{chengdiff,wei2025routing,li2025back}, we evaluate FID~\cite{heusel2017gans} and IS~\cite{salimans2016improved} using 50,000 generated images.
For latent-diffusion and pixel-space diffusion models, we follow their respective evaluation protocols, using CFG values of 1.0 and 1.5 for LDM-based methods, and the CFG interval protocol for JiTMoE.
Additionally, we assessed the model’s performance through visual inspection of samples generated from class conditions.

\subsection{Main Results}

\begin{figure}[ht]
    \centering
    \includegraphics[width=\linewidth]{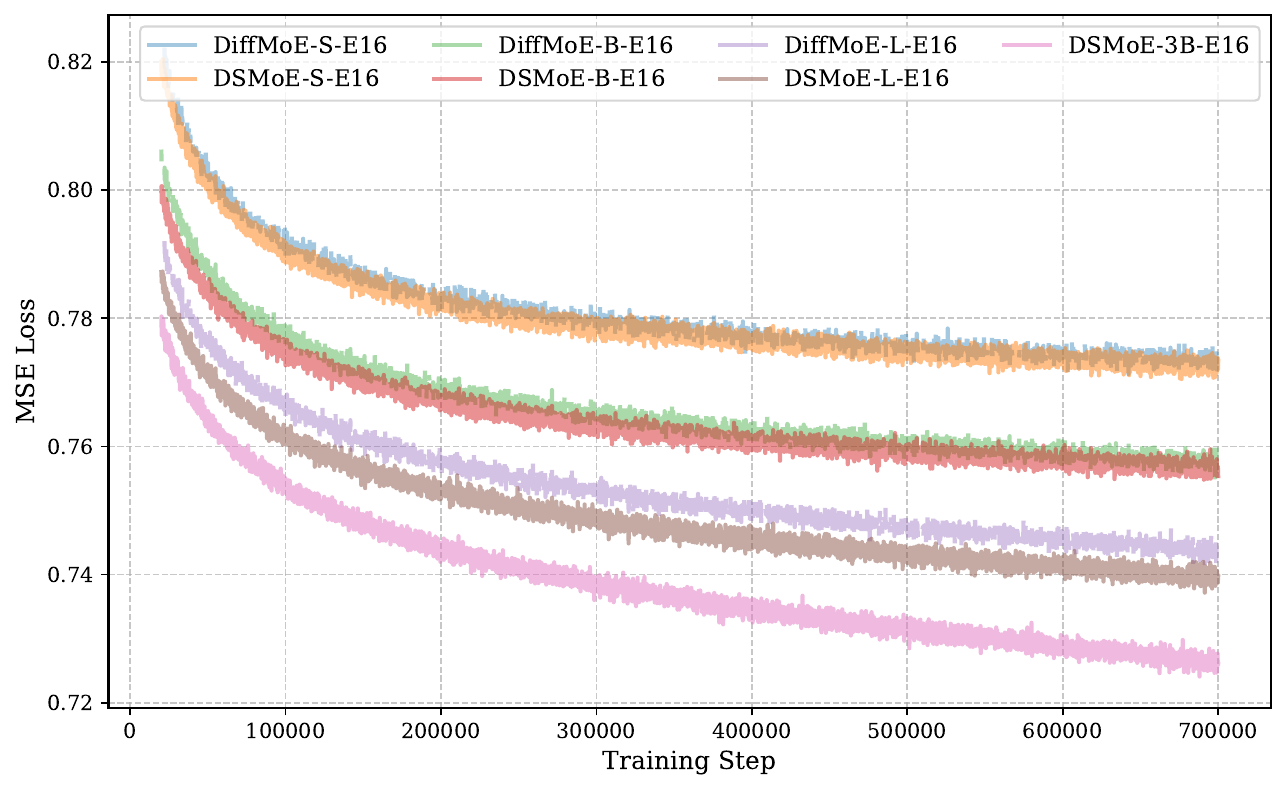}
    \caption{Training MSE loss curves of DiffMoE and DSMoE models. DSMoE, equipped with improved MoE designs, consistently achieves lower diffusion MSE losses than the baselines. Moreover, this performance gap further widens as the model scales up.}
    \label{fig:mse_loss_scaling}
\end{figure}

Table~\ref{tab:compared_diffmoe} compares DSMoE with DiffMoE~\cite{chengdiff} under the latent diffusion RF setup. Prior work typically uses a small number of experts (\textit{e.g.}, 8, 14, or 16) with wide intermediates (\textit{e.g.}, hidden\_size$\times$4), our design increases the expert count (\textit{e.g.}, 48) while shrinking intermediate width (\textit{e.g.}, hidden\_size$\times$0.3), which simultaneously lowers activated parameters and improves quality. At the B scale, DSMoE-B-E48 improves FID from 20.83 to 19.46 (CFG=1.0) and from 4.87 to 4.27 (CFG=1.5) while using roughly half the total parameters. At the L scale, DSMoE-L-E48 attains 9.19/2.55 FID versus DiffMoE’s 11.16/2.84 with a 44\% reduction in total parameters. The 3B model further pushes FID to 7.52/2.38 without increasing activated parameters relative to L. As shown in Figure~\ref{fig:mse_loss_scaling}, we plot training MSE loss for matched DiffMoE and DSMoE models, showing that our design not only converges faster but also attains consistently lower losses as scaling the model size, indicating improved optimization stability.

Table~\ref{tab:compared_promoe} benchmarks DSMoE against Dense-DiT~\cite{peebles2023scalable} and ProMoE~\cite{wei2025routing}. DSMoE consistently outperforms both while keeping activated parameters close to the dense baselines. For instance, DSMoE-L-E48 improves over ProMoE-L by 1.74 (CFG=1.0) and 0.07 (CFG=1.5) FID despite using fewer total parameters (1.11B vs. 1.06B activated 436M vs. 458M).

\begin{figure}[ht]
\centering
    \begin{minipage}[t]{\columnwidth}
        \centering
        \includegraphics[width=\textwidth]{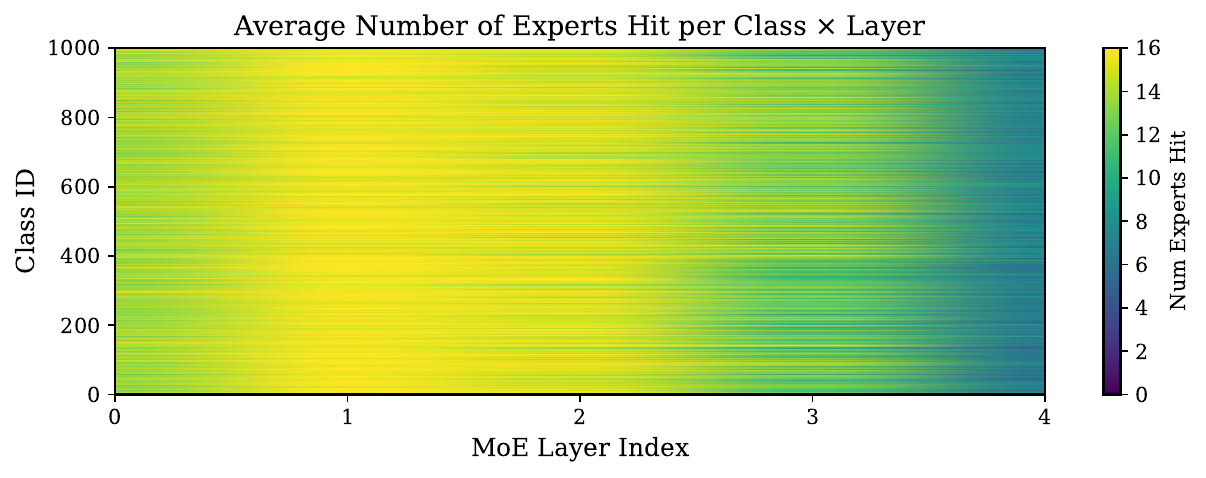}
    \end{minipage}
    \hfill 
    \begin{minipage}[t]{\columnwidth}
        \centering
        \includegraphics[width=\textwidth]{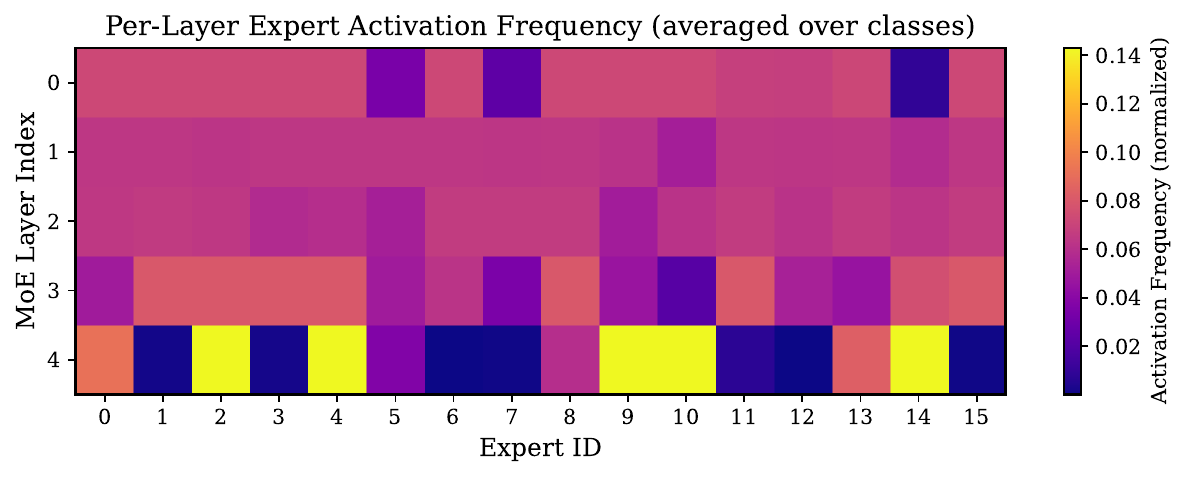}
    \end{minipage}
    \caption{
        Top: Average number of experts activated per class across MoE layers. Bottom: Per-layer expert activation frequency averaged over all classes.
    }
\label{fig:actived_experts}
\end{figure}

In Figure~\ref{fig:actived_experts}, we show expert usage analysis of our DSMoE-S-E16 by tracking the activations with 5,000 randomly sampled images. In Figure~\ref{fig:actived_experts} (Top), each row corresponds to a class, and each column to an MoE layer. The number of activated experts remains consistently high across classes and layers, indicating stable routing behavior. This indicates that:
1) the routing network does not collapse into a small subset of experts; 2) most experts are frequently utilized; and 3) load distribution remains balanced. 
Meanwhile, different experts exhibit distinct activation patterns across layers in Figure~\ref{fig:actived_experts} (Bottom). Several observations emerge: 1) different experts dominate at different layers, showing that specialization naturally emerges; 2) specialization is layer-dependent: certain experts are used more in early layers, while others are preferred in later layers. 
This pattern implies that experts learn functionally distinct roles, similar to how different convolutional filters specialize in edges, textures, or semantics. Moreover, in deeper layers (\textit{e.g.}, MoE layer 4), the activation patterns become more pronounced and unequal. This indicates that (1) the later layers capture increasingly abstract or higher-level features, and (2) token representations become more separable with respect to type, noise level, or semantic content.

\begin{figure}[ht]
    \centering
    \includegraphics[width=\linewidth]{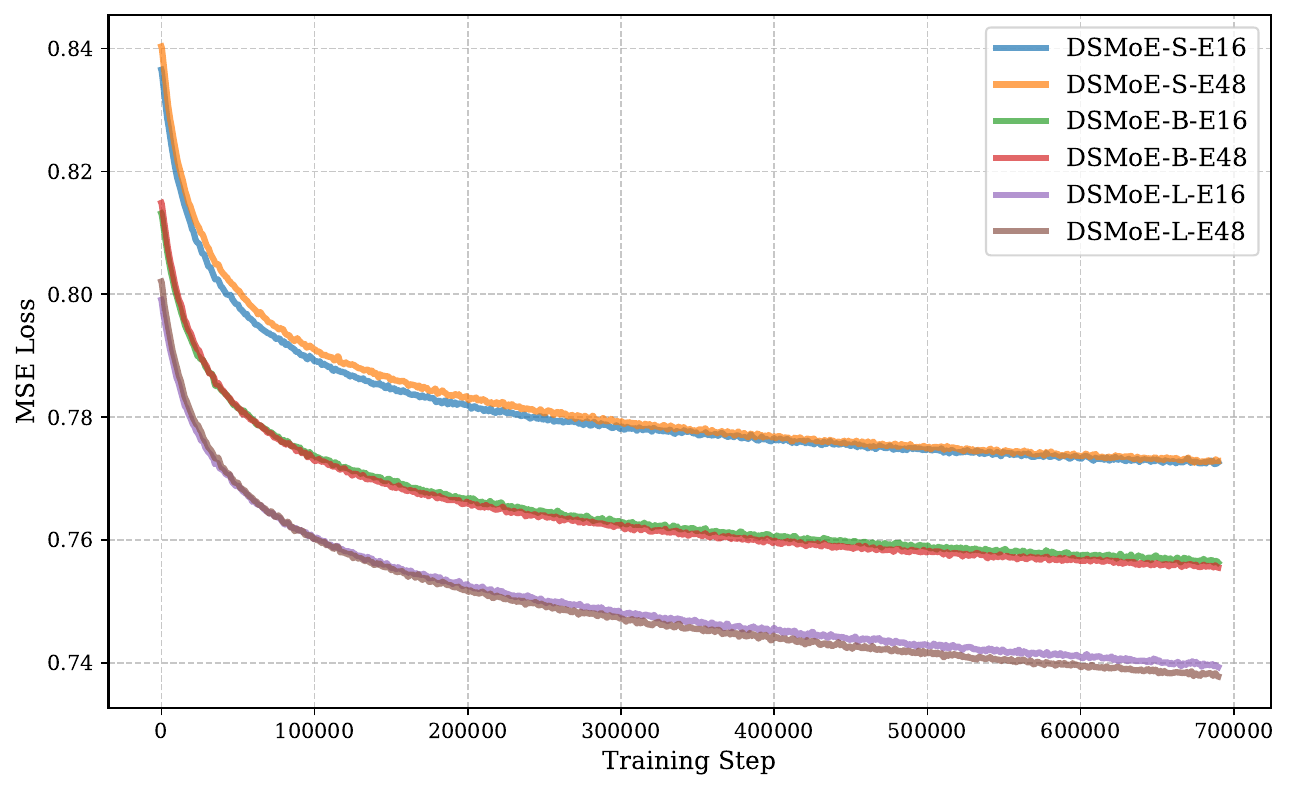}
    \caption{Training MSE loss curves of DSMoE with ``E16'' and ``E48''. Across all model sizes, E48 consistently achieves lower MSE losses than E16, indicating faster and more stable convergence.}
    \label{fig:e16-vs-e48-loss}
\end{figure}

As shown in Figure~\ref{fig:e16-vs-e48-loss}, increasing the number of experts, with a corresponding reduction in the FFN intermediate width to keep activated parameters constant, consistently yields lower MSE losses. This demonstrates faster and more stable training convergence.

Table~\ref{tab:compared_jit} extends the recipe to pixel-space diffusion JiT~\cite{li2025back}. JiTMoE-B/16-E16 slightly improves FID over JiT-B/16 (4.23 vs. 4.37) while dramatically boosting IS (245.53). On larger JiTMoE-L/16-E16, we also achieve performance comparable to JiT-L/16. 

Overall, the architectural changes (DeepSeek-style experts, narrowed intermediates, shared experts, and 2D RoPE) deliver consistent gains across latent and pixel diffusion models, with better parameter efficiency than routing-heavy baselines (\textit{e.g.}, DiffMoE and ProMoE).

\begin{table*}[ht]
  \centering
\caption{Quantitative comparison with JiT~\cite{li2025back} on various model sizes. All models are trained on ImageNet 256x256 with 200 epochs (batch size 1024). * refers to our reproduced performence.
}
  {
    \begin{tabular}{lccccccc}\toprule
    \multirow{2}{*}{Model} &
    \#Activated  & \#Total 
    & \multicolumn{2}{c}{w/ CFG interval}
    \\ \cmidrule(lr){4-5} 
    & Params. & Params. & FID ($\downarrow$) & IS ($\uparrow$) \\  \midrule
    JiT-B/16 & 131M & 131M & 4.37 (4.81*) & - (222.32*)  \\
    \rowcolor{rowgray}
    JiTMoE-B/16-E16 & 133M & 369M & 4.23 &  245.53 \\
    \midrule
    JiT-L/16 & 459M & 459M & 2.79 (3.19*) & - (309.72*)  \\
    \rowcolor{rowgray}
    JiTMoE-L/16-E16 & 465M & 1.306B & 3.10 & 311.34 
    \\
    \bottomrule
    \end{tabular}%
    }
  \label{tab:compared_jit}%
\end{table*}%

\begin{table*}[ht]
  \centering
\caption{Ablation study on positional encoding (PE) methods. All models are trained on ImageNet 256x256 with 700K steps. 
}
  {
    \begin{tabular}{lccccccc}\toprule
    \multirow{2}{*}{Model} & 
    \multirow{2}{*}{PE method} 
    & \multicolumn{2}{c}{CFG = 1.0}
    & \multicolumn{2}{c}{CFG = 1.5} 
    \\ \cmidrule(lr){3-4} \cmidrule(lr){5-6}
    & & FID ($\downarrow$) & IS ($\uparrow$) & FID ($\downarrow$) & IS ($\uparrow$)
    \\ \midrule
    \multirow{3}{*}{DSMoE-S-E16} & APE & 45.13 & 33.27 & 18.10 & 82.37 \\
     & RoPE (1D) & 44.75 & 33.64 & 18.12 & 83.13 \\
     & RoPE (2D) & 39.84 & 38.63 & 14.53 & 97.55 \\
    \bottomrule
    \end{tabular}%
    }
  \label{tab:ablation_pe}%
\end{table*}%

\subsection{Ablation Study}

\begin{figure}[ht]
    \centering
    \includegraphics[width=\linewidth]{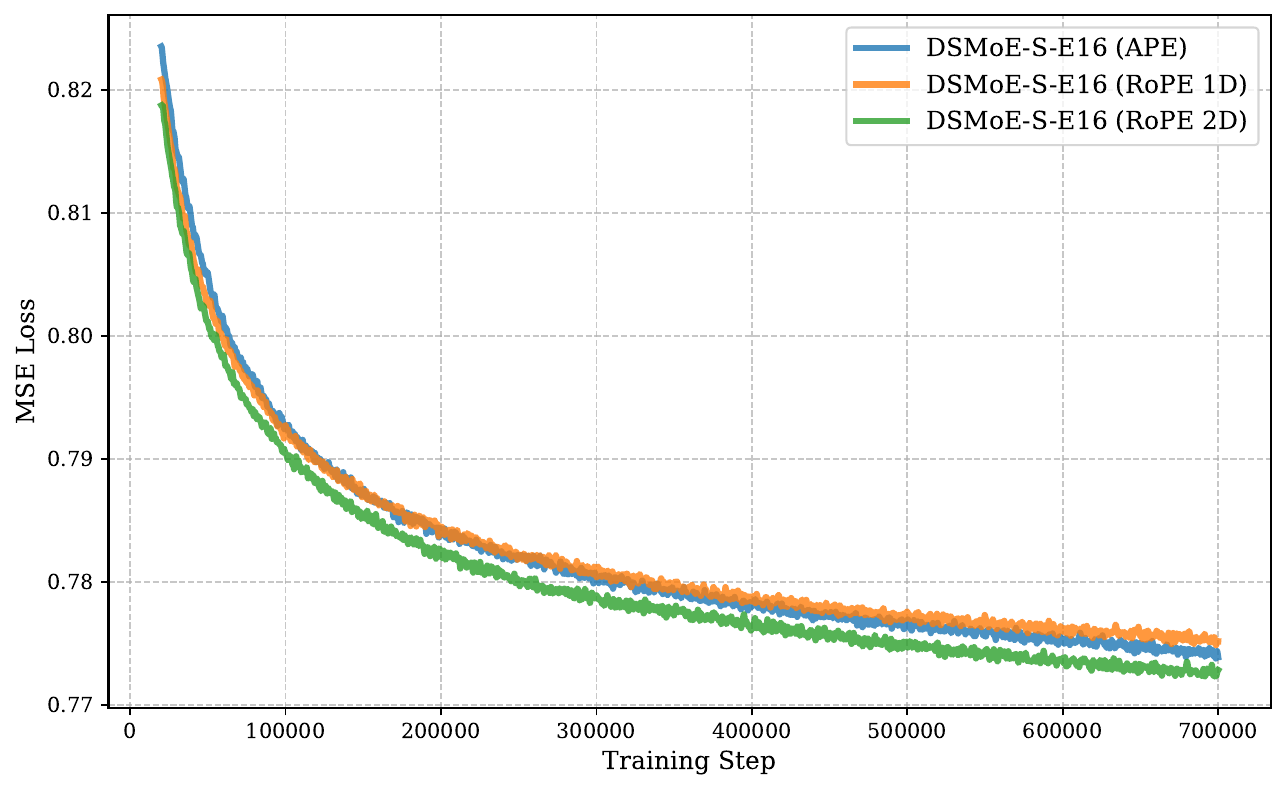}
    \caption{Training MSE loss curves of DSMoE with different position encoding methods.}
    \label{fig:mse_loss_pe}
\end{figure}

\noindent\textbf{On Positional Encoding.}
Table~\ref{tab:ablation_pe} demonstrates the effect of positional encoding on DSMoE-S-E16. Moving from absolute position embeddings (APE) to 1D RoPE yields a small improvement, but 2D RoPE delivers a large boost: FID drops from 45.13 to 39.84 (CFG=1.0) and from 18.10 to 14.53 (CFG=1.5), while IS improves from 33.27 to 38.63 (CFG=1.0) and from 82.37 to 97.55 (CFG=1.5). This validates that encoding row/column structure is critical for diffusion MoE training stability and quality. As shown in Figure~\ref{fig:mse_loss_pe}, we observe that 2D RoPE leads to faster training convergence, reflecting by consistently lower MSE losses than both APE and 1D RoPE.

\begin{figure}[ht]
    \centering
    \includegraphics[width=\linewidth]{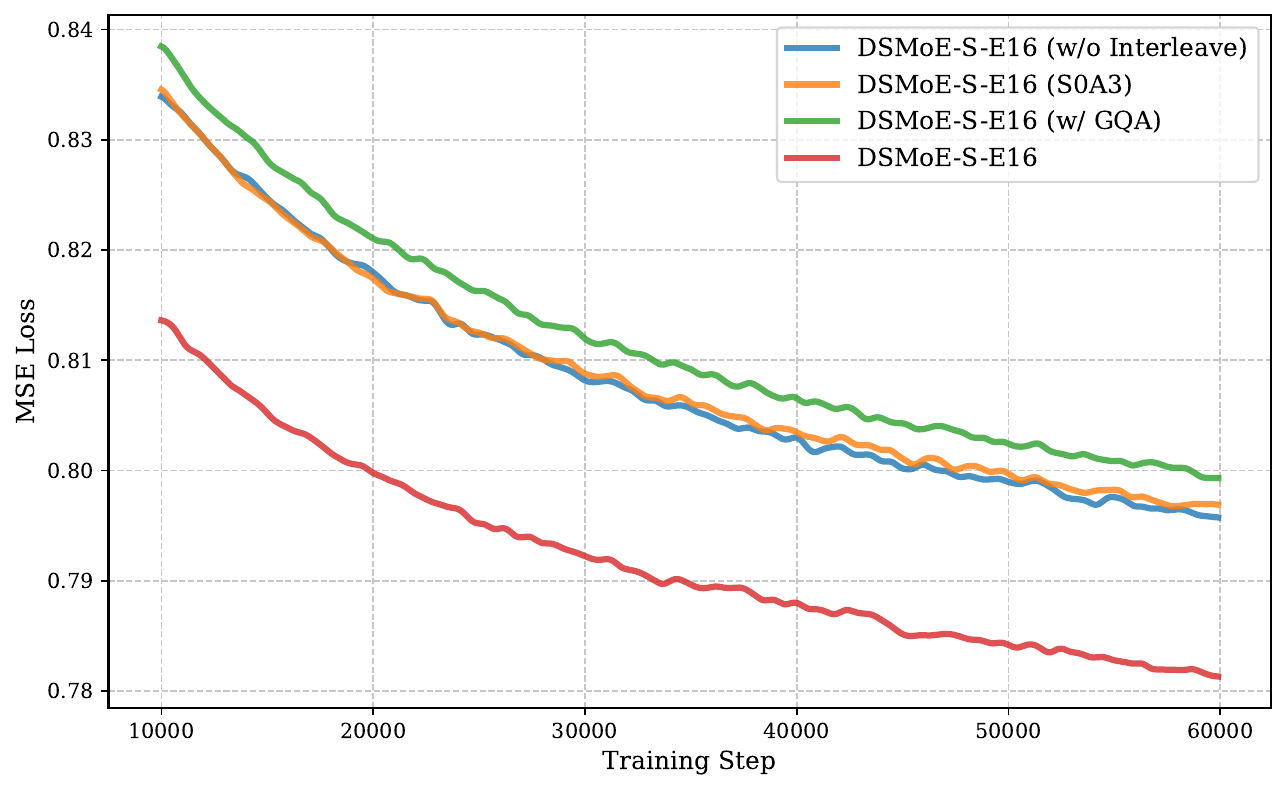}
    \caption{Ablations on replacing all FFN layers with MoE layer (\textit{i.e.}, w/o Interleave), applying GQA~\cite{ainslie2023gqa} (\textit{i.e.}, w/ GQA), and removing shared experts (\textit{i.e.}, S0A3).}
    \label{fig:ablation-all-gqa-s0}
\end{figure}

\noindent\textbf{On Shared Experts.} In DiffMoE~\cite{chengdiff}, the shared experts are not employed. We further investigate the role of shared experts by conducting an experiment where we remove them and instead increase the number of activated experts by one. In this manner, we keep the number of activated parameters unchanged. As shown in Figure~\ref{fig:ablation-all-gqa-s0}, this variant (\textit{i.e.}, S0A3) converges more slowly. This suggests that the shared expert plays an important stabilizing role: it provides a consistent fallback representation across tokens and timesteps, reducing routing variance and improving optimization smoothness. Consequently, models with a shared expert tend to train more reliably and converge faster.

\noindent\textbf{On Interleave of Using MoE Layers.} In previous Diffusion MoE models~\cite{chengdiff}, MoE layers are applied only to the even DiT blocks. We also evaluate a variant where all FFN layers are replaced with MoE layers. This modification substantially increases the total parameters from 92M to 158M. With 1 shared expert and 2 activated experts per token, the number of activated parameters rises to 44M (compared with 33M in our default setting). However, this configuration yields slight slower convergence, as shown in Figure~\ref{fig:ablation-all-gqa-s0} (\textit{i.e.}, w/o Interleave). We speculate that applying MoE to every FFN layer removes the alternation between dense and sparse computations, which normally provides a balance between stable global representations and specialized expert behaviors. In other words, interleaving dense and MoE layers helps maintain feature smoothness, reduces routing variance, and stabilizes optimization.

\noindent\textbf{On Efficiency of the Attention Mechanism.} GQA~\cite{ainslie2023gqa} and MLA~\cite{liu2024deepseekv2} methods have shown strong potential for compressing KV caches while preserving inference performance. We apply GQA to the DiT attention module and observe a slight performance degradation, as shown in Figure~\ref{fig:ablation-all-gqa-s0} (\textit{i.e.}, w/ GQA). When applying GQA, many heads share the same K/V projections, effectively: 1) reducing the diversity of attention patterns, 2) weakening the model’s ability to capture fine-grained spatial structures. These are harmful for the iterative denoising process.

\subsection{Visual Generation Results} 

\begin{figure}[ht]
    \centering
    \includegraphics[width=\linewidth]{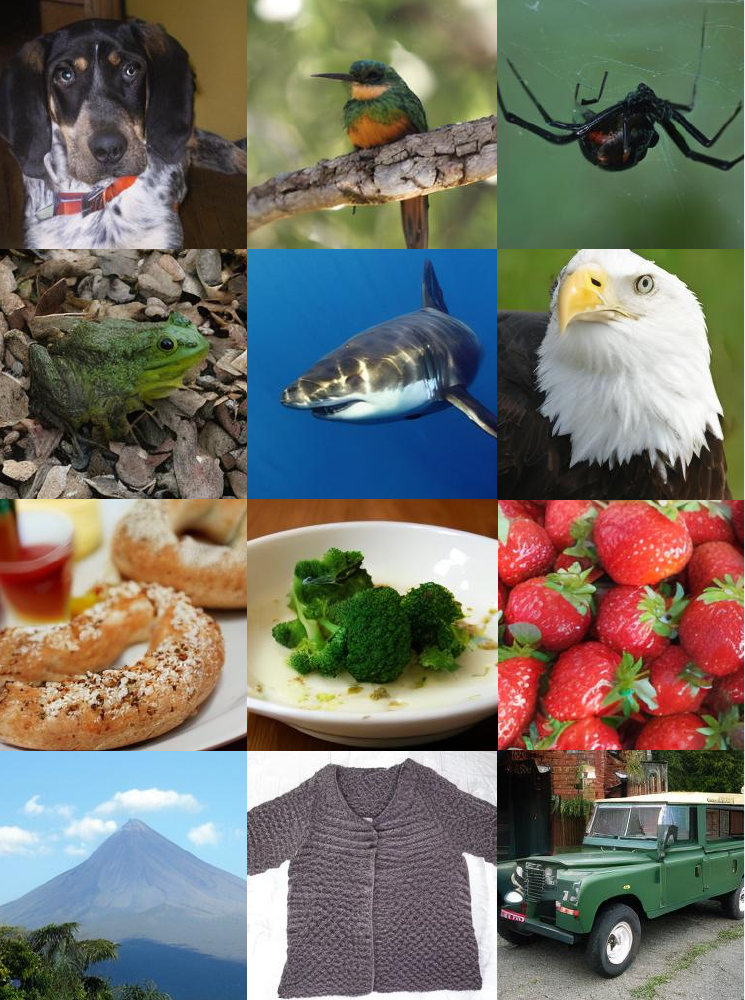}
    \caption{Class-conditional Generation with DSMoE-3B-E16. Uncurated 256$\times$256 samples CFG scale=1.5.}
    \label{fig:vis-dsmoe-3b}
\end{figure}

\begin{figure}[ht]
    \centering
    \includegraphics[width=\linewidth]{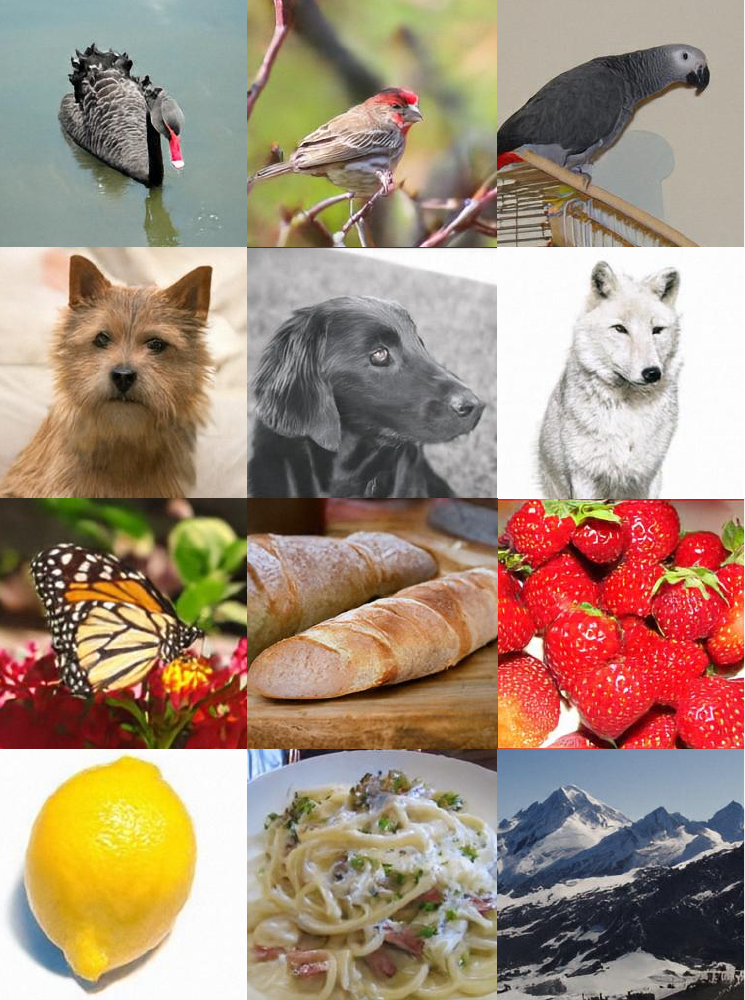}
    \caption{Class-conditional Generation with JiTMoE-L/16-E16. Uncurated 256$\times$256 samples with CFG interval.}
    \label{fig:vis-jitmoe-l}
\end{figure}

To demonstrate the generation capabilities of our model, we showcase diverse images sampled from DSMoE-3B-E16
and JiTMoE-L/16-E16, conditioned on ImageNet class labels. These visualizations illustrate the model’s ability to generate high-quality, class-specific images. See Figure~\ref{fig:vis-dsmoe-3b} and Figure~\ref{fig:vis-jitmoe-l}.

%% file: sections/5-conclusion.tex
\section{Conclusion}
We presented a practical recipe for diffusion MoE models that emphasizes architecture over routing. DeepSeek-style experts with narrowed intermediates, shared experts, and 2D RoPE yield DSMoE and JiTMoE models that outperform dense and prior MoE baselines on ImageNet 256$\times$256 under Rectified Flow, with lower activated parameters and improved training curves. The same design scales smoothly across model sizes and pixel/latent diffusion settings.

Future work includes combining these architectural choices with stronger routing strategies, expanding to text-to-image and video tasks, and exploring the reinforcement learning on the diffusion MoE models.

%% file: sections/appendix.tex
\appendix
\section{Detailed Configurations}

For training DSMoE models, we follow DiffMoE~\cite{chengdiff} use horizontal flips as the only data augmentation. We train all models with AdamW~\cite{loshchilov2017decoupled}. 
We use a constant learning rate of 1$\times$10$^{-4}$, no weight decay and a fixed global batch size of 512. 
We also maintain an exponential moving average (EMA) of DSMoE and baseline model weights over training with a decay of 0.9999. All results reported use the EMA mode. We utilized NVIDIA H800 GPUs during training. 

For training JiTMoE models, we follow JiT~\cite{li2025back} and keep the main training and sampling procedures unchanged, as shown in Table~\ref{tab:jit-config}. For further details, refer to the original JiT paper.

\begin{table}[ht]
\centering
\caption{Configurations of experiments on JiT and JiTMoE models. Except for the MoE components, we keep the remaining architecture, training procedure, and sampling pipeline unchanged.}
\setlength{\tabcolsep}{5pt}
\resizebox{\columnwidth}{!}{%
\begin{tabular}{c|cc} 
\toprule
 & \textbf{B} & \textbf{L} \\
\rowcolor[gray]{0.9} \textbf{Architecture} & &  \\
depth & 12 & 24 \\
hidden dim & 768 & 1024 \\
heads & 12 & 16 \\
image size & \multicolumn{2}{c}{256} \\
patch size & \multicolumn{2}{c}{\texttt{image\_size} / 16} \\
bottleneck & \multicolumn{2}{c}{128 (B/L)} \\
dropout & \multicolumn{2}{c}{0 (B/L)} \\
in-context class tokens & \multicolumn{2}{c}{32 (if used)} \\
in-context start block & 4 & 8 \\
\rowcolor[gray]{0.9} \textbf{Training} & &  \\
epochs & \multicolumn{2}{c}{200} \\
warmup epochs & \multicolumn{2}{c}{5} \\
optimizer & \multicolumn{2}{c}{Adam, $\beta_1, \beta_2=0.9, 0.95$} \\
batch size & \multicolumn{2}{c}{1024} \\
learning rate & \multicolumn{2}{c}{2e-4} \\
learning rate schedule & \multicolumn{2}{c}{constant} \\
weight decay & \multicolumn{2}{c}{0} \\
ema decay & \multicolumn{2}{c}{\{0.9996, 0.9998, 0.9999\}} \\
\multirow{2}{*}{time sampler} & \multicolumn{2}{c}{$\text{logit}(t){\sim}\mathcal{N}(\mu, \sigma^2)$,} \\ & \multicolumn{2}{c}{$\mu=-0.8, \sigma=0.8$} \\
noise scale & \multicolumn{2}{c}{1.0 $\times$ \texttt{image\_size} / 256} \\
clip of $(1-t)$ in division & \multicolumn{2}{c}{0.05} \\
class token drop (for CFG) & \multicolumn{2}{c}{0.1} \\
\rowcolor[gray]{0.9} \textbf{Sampling} & &  \\
ODE solver & \multicolumn{2}{c}{Heun} \\
ODE steps & \multicolumn{2}{c}{50} \\
time steps & \multicolumn{2}{c}{linear in [0.0, 1.0]} \\
CFG scale sweep range & \multicolumn{2}{c}{[1.0, 4.0]} \\
CFG interval & \multicolumn{2}{c}{[0.1, 1] (if used)} \\
\bottomrule
\end{tabular}}
\label{tab:jit-config}
\end{table}